\documentclass{article}
\usepackage{spconf,amsmath,graphicx,color,cite}

\usepackage[algoruled]{algorithm2e}

\usepackage[table]{xcolor}
\usepackage{collcell}
\usepackage{hhline}
\usepackage{pgf}
\usepackage{multirow}
\usepackage{booktabs}
\usepackage{breqn} 
\def\colorModel{hsb} 

\newcommand\ColCell[1]{
    \pgfmathparse{#1<50?1:0}  
    \ifnum\pgfmathresult=0\relax\color{white}\fi
    
    \pgfmathsetmacro\compA{0} 
    \pgfmathsetmacro\compB{0} 
    \pgfmathsetmacro\compC{(1 - #1/100)}      
    \pgfmathparse{#1==0}
    \ifnum\pgfmathresult=1
        \edef\x{\noexpand\centering\noexpand\cellcolor[\colorModel]{\compA,\compB,\compC}}\x  
    \else
        \edef\x{\noexpand\centering\noexpand\cellcolor[\colorModel]{\compA,\compB,\compC}}\x #1
    \fi
} 
\newcolumntype{E}{>{\collectcell\ColCell}{r}<{\endcollectcell}}  

\def\x{{\mathbf x}}

\title{Lead2Gold: Towards exploiting the full potential of noisy transcriptions for speech recognition}
\name{$\text{Adrien Dufraux}^{1,2}$, $\text{Emmanuel Vincent}^{2}$, $\text{Awni Hannun}^{1}$, $\text{Armelle Brun}^{2}$, $\text{Matthijs Douze}^{1}$}
\address{${}^1$Facebook AI Research, ${}^2$Universit\'e de Lorraine, CNRS, Inria, LORIA, F-54000 Nancy, France}

\begin{document}

\maketitle

\begin{abstract}
The transcriptions used to train an Automatic Speech Recognition (ASR) system may contain errors.
Usually, either a quality control stage discards transcriptions with too many errors, or the noisy transcriptions are used as is.
We introduce Lead2Gold, a method to train an ASR system that exploits the full potential of noisy transcriptions.
Based on a noise model of transcription errors, Lead2Gold searches for better transcriptions of the training data with a beam search that takes this noise model into account.
The beam search is differentiable and does not require a forced alignment step, thus the whole system is trained end-to-end.
Lead2Gold can be viewed as a new loss function that can be used on top of any sequence-to-sequence deep neural network.
We conduct proof-of-concept experiments on noisy transcriptions generated from letter corruptions with different noise levels.
We show that Lead2Gold obtains a better ASR accuracy than a competitive baseline which does not account for the (artificially-introduced) transcription noise.

\end{abstract}

\section{Introduction}
\label{sec:intro}

Automatic Speech Recognition (ASR) systems are typically trained in a fully supervised fashion using paired data, i.e., speech utterances with the corresponding transcriptions.
The speech utterances can be either prepared or spontaneous.
Prepared speech involves well-constructed sentences that could be found in a written document or a prepared talk, whereas spontaneous speech is unprepared and results from people talking freely. 
Prepared speech datasets typically consist of read speech, such as the LibriSpeech \cite{panayotov2015librispeech} or the Wall Street Journal (WSJ) \cite{paul1992design} datasets. In this case, the ground truth transcriptions are directly available from the book or the newspaper that the speaker is reading.
Spontaneous speech is typically found in conversational speech datasets, such as the Fisher \cite{cieri2004fisher} or the Switchboard \cite{godfrey1992switchboard} datasets.
In this case, the speech signal must be manually transcribed.

One the one hand, large read speech datasets can easily be collected for a small cost but they are not representative of the real test conditions encountered in commercial applications. As a result, models trained on these datasets achieve limited performance in real conditions. On the other hand, spontaneous speech datasets better match real test conditions but they are much more costly to acquire because of the manual transcription process. The transcriptions commonly used contain errors, which lead to performance degradation. The transcribers may provide higher quality transcriptions but at a much higher cost.

A few studies have sought to reduce the transcription cost. One approach is to perform semi-supervised learning on both transcribed and untranscribed data \cite{vesely2013semi,karita2018semi}. Another approach is to use active learning to selectively transcribe some utterances \cite{huang2016active}. The two approaches can be combined \cite{drugman2019active,Yu2010active}. Several studies have also considered transfer learning from related languages \cite{wang2015transfer}. These approaches help reducing the cost, but the amount of knowledge that can be inferred from untranscribed data is intrinsically lower.

In this paper, we explore the middle ground where the training data are neither accurately transcribed nor untranscribed but a not-so-expensive ``noisy'' transcription is available instead. We propose a new method called \textit{Lead2Gold} that learns an end-to-end ASR model given a noise model and a single noisy transcription per utterance. We exploit all available information as opposed to discarding overly noisy transcriptions.
To do so, we adapt the \emph{Auto Segmentation Criterion} (ASG) loss \cite{collobert2016wav2letter} to account for several possible transcriptions.
The probability of every possible alignment for a given transcription is obtained from the current ASR model and from a noise model which quantifies how likely it has been mistranscribed into the provided noisy transcription.
Because the computation of this loss is intractable, we use a differentiable beam search algorithm that samples only the best alignments of the best transcriptions. 

Noise models have been used in the field of image classification. 
They can either be learned in advance~\cite{patrini2017making} or jointly with the rest of the model with an extra layer~\cite{Label-Noise, sukhbaatar2014training, jindal2016learning, goldberger2017training}. Prior work has also used a noise model conditioned on the input features~\cite{xiao2015learning, menon2016learning}. However, these models cannot be directly applied to ASR as they do not handle sequential inputs and arbitrary-length outputs.
Fewer studies have considered noise models for ASR. In \cite{spellingCorrection}, a spelling correction model was used for the distinct goal of correcting the output of an already trained end-to-end ASR system. In \cite{hasegawa2017asr}, the phonetic sequence corresponding to each training utterance was inferred from several noisy transcriptions made by non-native transcribers using a misperception model, and subsequently used to train a conventional ASR model. Lead2Gold stands apart as it considers a sequence-level loss, it jointly learns the ASR model and the transcription graph (hence resulting in better transcriptions of the training data than using the noise model alone), and it requires a single noisy transcription per utterance.

This paper is a proof-of-concept for the proposed sequence-level noise model and training algorithm. As such, we adopt a controlled experimental protocol where we choose the noise model and generate noisy datasets accordingly. We present the algorithm in Section~\ref{sec:method}, then the experimental protocol and the experimental results in Sections~\ref{sec:preliminary} and \ref{sec:expes}, respectively. We conclude in Section~\ref{sec:conclu}.

\section{Searching for and learning from better transcriptions}
\label{sec:method}

Let us consider an utterance consisting of a feature sequence $X = [X_1,...,X_T]$ over $T$ frames and the corresponding provided transcription $Y = [y_1,...,y_L]$ which is a sequence of $L$ tokens.
An alignment of a transcription over the $T$ frames is denoted as $\pi = [\pi_1,...,\pi_T]$, where $\pi_t$ is the token in frame $t$.
Given an utterance $X$, the acoustic model outputs token scores $f_{\pi_t}(X)$ for each frame $t$.

Our method relies on the sequence-discriminative ASG loss \cite{collobert2016wav2letter}, that is an alternative to the \emph{Connectionist Temporal Classification} (CTC) loss \cite{graves2006connectionist} often used to train end-to-end systems. ASG incorporates bigram frame-level transitions, does not contain an optional null output, and is globally normalized.
As such, it is similar to the lattice-free Maximum Mutual Information (LF-MMI) loss also commonly used in ASR \cite{povey2016purely}.

With this approach, a forced alignment step is not needed before training since all possible alignments of the provided transcription are taken into account. The score of every individual alignment is defined as
\begin{equation}
    s(\pi|X) = \sum^T_{t=1}(f_{\pi_t}(X) + g(\pi_t|\pi_{t-1}))
\end{equation}
where $g(i|j)$ are the transition scores learned jointly with the acoustic model. The ASG loss to be minimized is defined as
\begin{align}
\text{ASG}(Y) &= - \log p(Y|X) \\ 
&= - \underset{\pi \in G_{ASG}(Y)}{\text{logadd}} s(\pi|X) + \underset{\pi \in G_{full}}{\text{logadd}} s(\pi|X) \\
&= - S_{ASG}(Y) + Z
\end{align}
where the logadd operation is defined as $\text{logadd}(a+b)=\log(e^a + e^b)$.
$G_{ASG}(Y)$ is the graph comprising all possible alignments of the transcription $Y$ over $T$ frames. Contrary to \cite{collobert2016wav2letter}, we use only one additional token to model the repetition of the previous letter and do not model the repetition of 3 consecutive letters.
The graph $G_{full}$ contains all possible alignments of all possible transcriptions.
The first term $S_{ASG}(Y)$ promotes all the paths in $G_{ASG}(Y)$, while the second term $Z$ is a normalization term.
Both terms are efficiently computed with a dynamic programming algorithm.

\subsection{Noise model}
\label{sec:noise_model}
In this work, we consider the case when the provided transcriptions are corrupted.
For each utterance, we denote by $\tilde{Y} = [\tilde{y}_1,...,\tilde{y}_{L_1}]$ the provided noisy transcription, and by $Y^* = [y_1^*,...,y_{L_2}^*]$ the unknown correct (a.k.a.\ clean) transcription.
We consider a noise model $p(\tilde{Y} | Y^*)$ which provides the likelihood of a noisy transcription given the clean one.
Note that the noise model conditions on the clean transcription only and does not depend on the utterance $X$.

In the following, as a proof-of-concept, we consider a simple noise model comprising token-level corruptions only.
Clean transcriptions can be transformed into noisy transcriptions by substituting tokens, deleting clean tokens, or inserting new tokens into the resulting noisy transcription.
For simplicity, we do not allow the deletion or insertion of two consecutive tokens.
We also assume that insertions, deletions and substitutions are independent of each other.

To model this behavior we add a void token $\emptyset$.
The  probability of deleting a clean token $y_i^*$ is given by $p(\emptyset | y_i^*)$ and $p(\tilde{y}_i | \emptyset)$ is the probability of inserting a noisy token $\tilde{y}_i$.
We interleave $\emptyset$ between all the tokens in $\tilde{Y}$ and $Y^*$ so that 
$\tilde{Y}$ becomes $[\emptyset,\tilde{y}_1,\emptyset,...,\emptyset,\tilde{y}_{L_1},\emptyset]$ and $Y^*$ becomes $[\emptyset,y_1^*,\emptyset,...,\emptyset,y_{L_2}^*,\emptyset]$.
A valid alignment between $\tilde{Y}$ and $Y^*$  is denoted by $a = (a^*,\tilde{a})$ where $a^*$ and $\tilde{a}$ contain $L_a$ tokens.
\newcommand{\allalign}{A_{Y^*}^{\tilde{Y}}}
\newcommand{\logadd}[1]{\underset{#1}{\textrm{logadd}}}
We denote by $\allalign$ the set of all possible alignments between $\tilde{Y}$ and $Y^*$. 

The noise model can now be computed as follows:
\begin{align}
 \log p(\tilde{Y} | Y^*) &= \log \sum_{a \in A_{Y^*}^{\tilde{Y}}}  p(\tilde{Y} | Y^*, a) \\
                &= \log \sum_{a \in \allalign} \prod_{i=1}^{L_a} p(\tilde{a}_i | a^*_i) \\
                &= \logadd{a \in \allalign} \sum_{i=1}^{L_a} \log p(\tilde{a}_i | a^*_i).
\label{eq:noise_model_alignment}
\end{align}
\newcommand{\AllThetas}{\mathcal{Y}_{\tilde{Y}}}
We denote by $\AllThetas$ the set of all clean transcriptions $Y^*$ that can lead to $\tilde{Y}$ under this noise model: 
\begin{equation}
\AllThetas = \{Y^* \mid p(\tilde{Y} | Y^*) > 0 \}
\end{equation}

\noindent
\textbf{Substitutions only:}
We also consider the simplified case where only substitutions are allowed. In this case, only one alignment $a$ is possible between $\tilde{Y}$ and $Y^*$, the length of the alignment is the same as the clean and noisy transcriptions ($L_a\!=\!L_1\!=\!L_2$), and expression \eqref{eq:noise_model_alignment} reduces to
\begin{equation}
\log p(\tilde{Y} | Y^*) = \sum_{i=1}^{L_a} \log p(\tilde{y}_i | y_i^*).
\end{equation}

\subsection{Noise-aware ASG loss}
We now reformulate the ASG loss to incorporate our model of transcription noise. The resulting noise-aware ASG loss can be computed as follows:
\begin{align}
&\text{ASG}(\tilde{Y}) = - \log p(\tilde{Y}|X) \\ 
& = - \log \sum_{Y^* \in  \AllThetas} p(\tilde{Y} | Y^*) p(Y^*|X) \\
& =  - \logadd{Y^* \in  \AllThetas} \left [ \log p(\tilde{Y} | Y^*) + \log p(Y^*|X) \right ] \\
& =  - \logadd{Y^* \in  \AllThetas} \left [ \log p(\tilde{Y} | Y^*) - \text{ASG}(Y^*) \right ] \\
& =  - \logadd{Y^* } \left [ \log p(\tilde{Y} | Y^*)  + \logadd{\pi^* \in G_{ASG}(Y^*)} s(\pi^*|X) - Z \right ] \\
& =  - \logadd{Y^*} \left [ \log p(\tilde{Y} | Y^*)  + \underset{\pi^* }{\text{logadd}} \ s(\pi^*|X) \right ] + Z \\
& =  - \logadd{Y^*, \pi^*}
    \left [ s(\pi^*|X) + \log p(\tilde{Y} | Y^*) \right ] + Z \\
& = - S_{L2G}(\tilde{Y}) + Z.
\end{align}
The normalization term $Z$ is straightforward to compute with dynamic programming as it is unchanged with respect to the conventional ASG loss.

Including the noise model from Section~\ref{sec:noise_model} in $S_{L2G}(\tilde{Y})$, we obtain:
\begin{align}
S_{L2G}(\tilde{Y})
& = \logadd{Y^*,\pi^*}
    \left [ s(\pi^*|X) + \logadd{a\in \allalign} \sum_{i=1}^{L_a} \log p(\tilde{a}_i | a^*_i) \right ] \\
& = \logadd{Y^*,\pi^*, a}\,\,\left[
    s(\pi^*|X) + \sum_{i=1}^{L_a} \log p(\tilde{a}_i | a^*_i) 
    \right] \\
& = \logadd{(Y^*,\pi^*,a) \in G_{L2G(\tilde{Y})}} \ s(Y^*,\pi^*,a | X)
\label{eq:l2g_numerator}
\end{align}
where $G_{L2G(\tilde{Y})}$ is the graph comprising all the alignments $\pi^*$ over $T$ frames of all possible clean transcriptions $Y^*$ corresponding to the provided noisy transcription $\tilde{Y}$.
Given the combinatorially large number of paths in $G_{L2G(\tilde{Y})}$, $S_{L2G}(\tilde{Y})$ cannot be computed exactly.

\subsection{Approximation of the loss}
\begin{algorithm}[t]
\DontPrintSemicolon
\SetKwFunction{Add}{AddHyp}
\SetKwProg{Fn}{}{:}{}
\Fn{\Add{hyp, $\pi$, $\tilde{y}$, s}}{
         Create a new hyp which adds $\pi$ to hyp with cursor to $\tilde{y}$ and a score $s$, only if it leads to a valid path.\;
  }\;
  
Declare an empty hyp with cursor at $\tilde{y_1}$ and score to 0\;
\For{every frame $t$}{
    \For{every hyp being expanded}{
        hyp has the cursor $\tilde{y}_i$\ and a score $s_{prev}$\;
        \For{every token $\pi^*_t$}{
             $s = s_{prev} + f_{\pi^*_t}(X) + g(\pi^*_t|\pi^*_{t-1})$\;
             \eIf{$\pi^*_t = \pi^*_{t-1}$}{
                \Add(hyp, $\pi^*_t$, $\tilde{y}_i$, $s$)\;
             }{
                \textbf{Deletion of $\pi^*_{t-1}$:}\;
                \ \ $s_{t} = s + \alpha \log p(\emptyset|\pi^*_{t-1})$\;
                \ \ \Add(hyp, $\pi^*_t$, $\tilde{y}_i$, $s_{t}$)\;
                
                \textbf{Substitution of $\pi^*_{t-1}$ to $\tilde{y}_i$:}\;
                \ \ $s_{t} = s + \alpha \log p(\tilde{y}_i|\pi^*_{t-1})$\;
                \ \ \Add(hyp, $\pi^*_t$, $\tilde{y}_{i+1}$, $s_{t}$)\;
                
                \textbf{Insertion of $\tilde{y}_i$ and}\;
                \textbf{Substitution of $\pi^*_{t-1}$ to $\tilde{y}_{i+1}$:}\;
                \ \  $s_{t} = s + \alpha \log  p(\tilde{y}_i|\emptyset) p(\tilde{y}_{i+1}|\pi^*_{t-1})$\;
                \ \ \Add(hyp, $\pi^*_t$, $\tilde{y}_{i+2}$, $s_{t}$)\;
            }
        }
    }
    Merge hypotheses with same cursor and same $\pi^*_t$.\;
    Keep the $N$ best hypotheses.\;
}
Add noise score for the last token, logadd all score hypotheses and return the resulting score.\;

 \caption{Forward pass for the differentiable beam search.}
 \label{algo_beam}
\end{algorithm}   

To obtain an approximation of $S_{L2G}(\tilde{Y})$ we take into account only a subset of all possible paths in $G_{L2G(\tilde{Y})}$, that we obtain with a beam search algorithm.
The algorithm is inspired from ~\cite{collobert2019fully}, except that the way we choose the paths differs drastically.
$s(Y^*,\pi^*,a | X)$ can be computed as a sum over the $T$ frames and Algorithm~\ref{algo_beam} describes how we expand the hypotheses with a corresponding score at each frame.
The noise score is added once an entire letter is emitted.
The function \textit{AddHyp} prevents from adding an invalid hypothesis, i.e., the considered hypothesis must lead to a $Y^*$ that belongs to $\AllThetas$.
After each processed frame, we merge the hypotheses that have the same cursor on the noisy transcription and that are processing the same token.
Then we keep only the $N$ best hypotheses where $N$ denotes the beamsize. For a large enough $N$, we are able to compute a good approximation of $G_{L2G(\tilde{Y})}$.
We denote by $L2G(\tilde{Y})$ the loss obtained with this approximation.

To train a model with this loss, it has to be differentiable.
Once we choose the hypotheses in the forward pass, the backward pass doesn't change from ~\cite{collobert2019fully}.

During the forward pass, we add the acoustic scores with the log-probabilities of the noise model.
To make our beam search work, we have to balance the contribution of these two terms with a \emph{noise model weight} $\alpha$ akin to the language model weight in conventional ASR decoding.
This parameter is tuned manually on a validation set.

\subsection{Performing the transcription}

To perform ASR on the test set, we apply the standard transcription procedure, as if there was no corruption at training time.
The acoustic model was trained to model $p(Y^* | X)$ so given an utterance X, the model will output clean token probabilities for each frame.
We can either decode a transcription with a Viterbi algorithm or use a standard beam search with a LM.

\section{Experimental setup and preliminary observations}
\label{sec:preliminary}

\subsection{Getting a noise model}

In this work, we do not create a new ASR dataset with noisy transcriptions. 
While it would be more realistic, it would also be costly to annotate and more difficult to evaluate and reproduce. 
Instead, we chose to corrupt an existing paired text and audio dataset with near-perfect transcriptions.
To generate realistic transcription noise, we study the incorrect predictions of a weak ASR model, which we assume resembles the errors made by human transcribers.

We use a grapheme-based ASR model. The output set includes the alphabet plus the apostrophe, the space, and one repetition token to model the repetition of 2 letters.
\\
\\
\noindent
\textbf{Weak ASR Model:}
To generate meaningful corruptions, we train a weak ASR model on the si284 subset of the WSJ dataset~\cite{paul1992design} (82 hours).
This model has the same architecture and follows the same recipe as the one used to evaluate our method (see Section~\ref{ssec:protocol}).
We call it a weak model because we do not allow it to converge completely.
We stop training at three different stages, namely when the Letter Error Rate (LER) on the nov93dev subset reaches 30, 20, or 10\%.
We denote these models as M30, M20, and M10, respectively. 
\\
\\
\noindent
\textbf{Letter substitution probabilities:}
We evaluate the weak ASR model on the nov93dev and nov92 subsets (836 utterances) and compare the greedy decoded transcriptions with the ground truth at the token level.
To do so, we minimize the letter edit distance between them and compute the frequency-based, empirical probabilities, $p(\tilde{y}| y^*)$, for substitutions, insertions ($y^*=\emptyset$), and deletions ($\tilde{y} = \emptyset$).
\\
\\
\noindent
\textbf{Generating corruptions:}
To generate more variations, we reduce or amplify the number of substitutions, insertions, and deletions by applying a multiplicative factor $f\in\{0.5, 1, 2\}$. 
We call the resulting noise models ``M* f*" in the general case or ``S M* f*" in the substitution-only case (ignoring insertions and deletions). 
For example, in ``S M10 f2", the letter substitution probabilities are from a WSJ model which reaches an LER of 10\% on nov93dev, with only substitutions, and the probabilities are calculated with a multiplicative factor of 2.
We consider 18 such models.

\begin{table*}[t]
 \scriptsize
 \addtolength{\tabcolsep}{-0pt}
\newcommand\items{29}  
\resizebox{\textwidth}{!}{
\noindent\begin{tabular}{cc|*{\items}{E}|}
\multicolumn{1}{c}{} &\multicolumn{1}{c}{} &\multicolumn{\items}{c}{\large Clean letters} \\ \hhline{~*\items{}}
\multicolumn{1}{c}{} & 
\multicolumn{1}{c}{} & 
\multicolumn{1}{r}{$|$} &
\multicolumn{1}{r}{'} & 
\multicolumn{1}{r}{a} & 
\multicolumn{1}{r}{b} & 
\multicolumn{1}{r}{c} & 
\multicolumn{1}{r}{d} & 
\multicolumn{1}{r}{e} & 
\multicolumn{1}{r}{f} & 
\multicolumn{1}{r}{g} & 
\multicolumn{1}{r}{h} & 
\multicolumn{1}{r}{i} & 
\multicolumn{1}{r}{j} & 
\multicolumn{1}{r}{k} & 
\multicolumn{1}{r}{l} & 
\multicolumn{1}{r}{m} & 
\multicolumn{1}{r}{n} & 
\multicolumn{1}{r}{o} & 
\multicolumn{1}{r}{p} & 
\multicolumn{1}{r}{q} & 
\multicolumn{1}{r}{r} & 
\multicolumn{1}{r}{s} & 
\multicolumn{1}{r}{t} & 
\multicolumn{1}{r}{u} & 
\multicolumn{1}{r}{v} & 
\multicolumn{1}{r}{w} & 
\multicolumn{1}{r}{x} & 
\multicolumn{1}{r}{y} & 
\multicolumn{1}{r}{z} & 
\multicolumn{1}{r}{$\emptyset$} \\ \hhline{~~*\items{-}}
\multirow{\items}{*}{\rotatebox{90}{\large Noisy letters}} 
& $|$& 89.4 & 1.8 & 0.6 & 0.9 & 1.0 & 1.0 & 0.8 & 0.3 & 0.9 & 0.7 & 1.0 & 0.8 & 0.9 & 0.8 & 1.0 & 0.7 & 0.5 & 1.0 & 2.0 & 0.4 & 0.8 & 0.7 & 0.9 & 0.4 & 1.0 & 0.6 & 0.7 & 7.1 & 0.3    \\ \hhline{~*\items{}}
& '& 0.0 & 31.8 & 0.1 &  0  &  0  &  0  & 0.1 &  0  &  0  &  0  & 0.1 &  0  &  0  &  0  &  0  & 0.0 & 0.0 &  0  &  0  &  0  &  0  & 0.0 &  0  &  0  &  0  &  0  &  0  &  0  &  0     \\ \hhline{~*\items{}}
& a& 0.4 & 2.5 & 75.9 & 0.1 & 0.2 & 0.5 & 2.3 & 2.2 & 0.6 & 0.4 & 3.0 &  0  & 0.6 & 0.7 & 0.4 & 0.6 & 2.4 & 0.9 &  0  & 0.4 & 0.2 & 0.5 & 4.2 & 0.4 & 0.7 &  0  & 0.6 & 1.4 & 0.1    \\ \hhline{~*\items{}}
& b& 0.0 &  0  & 0.1 & 82.9 &  0  & 0.2 &  0  & 0.1 & 0.1 & 0.5 & 0.0 &  0  & 0.2 &  0  & 0.4 & 0.0 &  0  & 1.5 &  0  & 0.0 &  0  &  0  & 0.1 & 1.6 & 0.3 &  0  & 0.2 &  0  & 0.0    \\ \hhline{~*\items{}}
& c& 0.1 & 0.4 & 0.1 & 0.1 & 85.3 & 0.1 & 0.2 & 0.3 & 1.6 & 0.3 & 0.2 &  0  & 6.6 & 0.1 & 0.2 & 0.0 & 0.2 & 1.0 & 6.1 & 0.1 & 0.8 & 0.3 & 0.3 & 0.1 &  0  & 1.2 & 0.2 & 1.4 & 0.0    \\ \hhline{~*\items{}}
& d& 0.1 &  0  & 0.2 & 1.0 &  0  & 67.1 & 0.4 & 0.3 & 2.1 & 1.2 & 0.1 & 3.9 &  0  & 0.1 & 0.3 & 0.4 & 0.1 & 0.1 &  0  & 0.2 & 0.2 & 1.5 & 0.3 & 1.0 & 0.4 &  0  & 0.5 &  0  & 0.1    \\ \hhline{~*\items{}}
& e& 0.4 & 4.0 & 4.6 & 1.0 & 0.8 & 1.6 & 77.7 & 0.7 & 1.2 & 0.9 & 2.9 &  0  & 1.5 & 1.0 & 1.0 & 0.5 & 3.0 & 0.2 &  0  & 0.6 & 0.5 & 1.4 & 4.2 & 1.2 & 0.7 &  0  & 3.8 & 7.1 & 0.2    \\ \hhline{~*\items{}}
& f&  0  &  0  & 0.2 &  0  &  0  & 0.1 & 0.2 & 83.5 & 0.1 & 0.9 & 0.1 &  0  &  0  & 0.1 &  0  & 0.0 & 0.0 & 0.4 &  0  & 0.1 & 0.1 & 0.2 & 0.3 & 1.3 & 0.5 & 0.6 & 0.2 &  0  & 0.0    \\ \hhline{~*\items{}}
& g& 0.1 &  0  & 0.1 & 0.5 & 0.3 & 1.2 & 0.2 & 0.1 & 71.4 & 0.0 & 0.1 & 3.1 & 0.9 & 0.1 & 0.2 & 0.1 &  0  & 0.3 & 1.0 & 0.1 & 0.0 & 0.2 & 0.3 & 0.1 & 0.1 &  0  & 0.4 &  0  & 0.1    \\ \hhline{~*\items{}}
& h& 0.1 &  0  & 0.2 & 0.8 & 0.0 & 0.3 & 0.2 & 0.4 & 0.1 & 73.8 & 0.2 & 1.6 & 0.4 & 0.3 & 0.1 & 0.2 & 0.3 & 0.1 & 1.0 & 0.2 & 0.1 & 0.2 & 0.5 & 0.8 & 0.3 &  0  & 0.2 &  0  & 0.1    \\ \hhline{~*\items{}}
& i& 0.4 & 4.0 & 3.3 & 0.4 & 0.3 & 0.4 & 1.7 &  0  & 0.3 & 0.3 & 78.2 &  0  &  0  & 0.2 & 0.5 & 0.2 & 1.1 & 0.2 &  0  & 0.2 & 0.2 & 0.2 & 3.0 & 0.8 & 0.2 &  0  & 2.9 &  0  & 0.1    \\ \hhline{~*\items{}}
& j&  0  &  0  &  0  &  0  &  0  & 0.0 & 0.0 &  0  & 0.5 &  0  & 0.0 & 79.8 &  0  &  0  &  0  &  0  &  0  &  0  &  0  & 0.0 &  0  & 0.0 & 0.1 &  0  &  0  &  0  &  0  &  0  &  0     \\ \hhline{~*\items{}}
& k& 0.0 & 0.7 & 0.1 &  0  & 0.6 & 0.0 & 0.0 & 0.1 & 0.3 & 0.2 & 0.0 &  0  & 68.9 & 0.1 &  0  & 0.0 &  0  & 0.3 & 3.0 &  0  & 0.0 & 0.3 & 0.1 &  0  &  0  &  0  & 0.1 &  0  &  0     \\ \hhline{~*\items{}}
& l& 0.1 & 1.4 & 0.3 & 0.1 &  0  & 0.3 & 0.3 &  0  & 0.1 & 0.4 & 0.1 &  0  &  0  & 83.9 & 1.1 & 0.2 & 0.6 & 0.2 &  0  &  0  & 0.1 & 0.2 & 0.6 & 0.3 & 2.8 &  0  & 0.2 & 1.4 & 0.0    \\ \hhline{~*\items{}}
& m& 0.1 &  0  & 0.1 & 0.5 & 0.0 & 0.4 & 0.0 &  0  &  0  & 0.1 & 0.1 &  0  &  0  & 0.4 & 81.3 & 0.6 & 0.1 & 0.1 &  0  & 0.2 & 0.0 & 0.0 & 0.1 & 0.7 & 0.4 &  0  & 0.1 &  0  &  0     \\ \hhline{~*\items{}}
& n& 0.2 & 0.7 & 0.2 & 0.1 & 0.2 & 1.2 & 0.4 & 0.8 & 0.5 & 0.5 & 0.2 &  0  &  0  & 0.4 & 4.7 & 89.1 & 0.2 & 0.3 &  0  & 0.2 & 0.0 & 0.5 & 0.4 &  0  & 0.1 &  0  & 0.5 &  0  & 0.1    \\ \hhline{~*\items{}}
& o& 0.2 & 1.4 & 2.7 & 0.3 & 0.1 & 0.8 & 1.2 & 0.5 & 0.4 & 0.1 & 1.0 & 1.6 & 0.2 & 1.8 & 0.4 & 0.3 & 81.3 & 0.3 &  0  & 0.7 & 0.1 & 0.1 & 4.8 & 1.0 & 1.3 &  0  & 0.2 & 2.9 & 0.1    \\ \hhline{~*\items{}}
& p& 0.0 &  0  & 0.0 & 1.3 & 0.8 &  0  & 0.1 & 0.5 & 0.1 & 0.2 & 0.1 &  0  & 0.4 & 0.2 &  0  &  0  & 0.0 & 86.7 &  0  & 0.0 & 0.0 & 0.1 & 0.1 & 0.3 & 0.1 &  0  & 0.1 &  0  &  0     \\ \hhline{~*\items{}}
& q& 0.0 &  0  & 0.0 &  0  & 0.0 &  0  &  0  &  0  &  0  & 0.0 &  0  &  0  &  0  &  0  &  0  &  0  &  0  & 0.1 & 72.7 & 0.0 &  0  &  0  & 0.1 &  0  &  0  &  0  &  0  &  0  &  0     \\ \hhline{~*\items{}}
& r& 0.0 &  0  & 0.3 & 0.1 & 0.0 & 0.2 & 0.2 & 0.3 & 0.1 & 0.1 & 0.1 &  0  &  0  & 0.2 & 0.4 & 0.0 & 0.1 & 0.2 & 1.0 & 90.5 & 0.1 & 0.0 & 0.4 & 0.1 & 0.5 &  0  & 0.2 &  0  & 0.0    \\ \hhline{~*\items{}}
& s& 0.2 &  0  & 0.2 &  0  & 2.6 & 0.6 & 0.6 & 1.0 & 0.4 & 0.3 & 0.2 & 0.8 &  0  & 0.1 & 0.1 & 0.2 & 0.2 &  0  &  0  & 0.0 & 90.4 & 0.2 & 0.3 &  0  & 0.1 & 2.9 & 0.6 & 32.9 & 0.1    \\ \hhline{~*\items{}}
& t& 0.3 & 0.4 & 0.4 & 1.2 & 1.3 & 5.2 & 0.6 & 1.6 & 1.0 & 1.1 & 0.5 & 3.9 & 5.7 & 0.6 & 0.4 & 0.9 & 0.2 & 1.7 & 4.0 & 0.2 & 0.6 & 82.9 & 0.6 & 0.7 & 0.3 & 0.6 & 0.9 &  0  & 0.2    \\ \hhline{~*\items{}}
& u& 0.1 & 0.4 & 0.6 & 0.1 & 0.2 & 0.0 & 0.4 & 0.1 & 0.1 & 0.2 & 0.8 &  0  &  0  & 0.3 & 0.3 & 0.1 & 0.8 & 0.1 &  0  & 0.2 & 0.0 & 0.1 & 64.0 & 0.3 & 0.7 &  0  & 0.2 &  0  & 0.0    \\ \hhline{~*\items{}}
& v& 0.1 &  0  & 0.0 & 1.4 &  0  & 0.2 & 0.0 & 0.5 & 0.1 & 0.2 &  0  &  0  &  0  & 0.1 & 0.3 & 0.0 & 0.0 & 0.2 &  0  & 0.1 &  0  & 0.1 & 0.1 & 77.5 & 0.1 &  0  &  0  &  0  &  0     \\ \hhline{~*\items{}}
& w& 0.0 &  0  & 0.0 & 0.2 & 0.0 & 0.2 & 0.0 & 0.2 & 0.1 & 0.0 & 0.0 &  0  &  0  & 0.5 & 0.2 & 0.1 & 0.2 & 0.1 &  0  & 0.1 & 0.0 & 0.0 & 0.4 & 0.7 & 68.5 &  0  & 0.2 &  0  & 0.0    \\ \hhline{~*\items{}}
& x& 0.0 &  0  &  0  &  0  & 0.0 &  0  &  0  &  0  &  0  & 0.0 &  0  &  0  & 0.4 &  0  &  0  &  0  & 0.0 &  0  &  0  &  0  & 0.3 & 0.0 &  0  &  0  &  0  & 91.9 &  0  &  0  &  0     \\ \hhline{~*\items{}}
& y&  0  &  0  & 0.1 &  0  & 0.2 & 0.8 & 0.6 &  0  & 1.0 & 0.2 & 0.6 &  0  &  0  & 0.0 & 0.1 & 0.1 & 0.1 &  0  &  0  & 0.0 & 0.1 & 0.2 & 0.2 & 0.1 &  0  &  0  & 74.8 & 1.4 & 0.0    \\ \hhline{~*\items{}}
& z&  0  &  0  &  0  &  0  &  0  & 0.0 &  0  &  0  &  0  &  0  &  0  &  0  &  0  &  0  &  0  &  0  &  0  &  0  &  0  &  0  & 0.0 &  0  &  0  &  0  &  0  &  0  &  0  & 25.7 &  0     \\ \hhline{~*\items{}}
& $\emptyset$& 7.6 & 50.5 & 9.5 & 6.7 & 5.6 & 17.5 & 11.9 & 6.4 & 16.7 & 16.8 & 10.4 & 4.7 & 13.2 & 8.0 & 6.7 & 5.7 & 8.6 & 4.4 & 9.1 & 5.4 & 5.1 & 10.1 & 13.9 & 10.8 & 21.0 & 2.3 & 12.5 & 18.6 & 98.5    \\ \hhline{~~*\items{-}}
\end{tabular}
} 
\caption{
\label{tab:letterconfusion}
The ``M20 f1" noise model probabilities. The ``$|$" is the space token. Row $\emptyset$ contains the deletion probabilities and column $\emptyset$ contains the insertion probabilities, and $p(\emptyset|\emptyset)$ is the probability of not inserting any new token into the transcription.}
\end{table*}

Table~\ref{tab:letterconfusion} shows an example of the ``M10 f2" noise model.
Interesting confusions are s/z, presumably because of the existence of both British and US spellings, or k/c because they often produce the same hard `k' sound.
In this setting, insertion of letters in the noisy transcription is unlikely.
This is because the model from which we get the errors tends to output shorter transcriptions than the ground truth.

\subsection{Generating noisy transcriptions}
We test our approach on the LibriSpeech corpus~\cite{panayotov2015librispeech}. 
For training we use the train-clean-100 subset which contains 100 hours of clean speech along with their ground truth transcriptions.
We use the smaller 100 hour subset because our approach is computationally intensive.
However, we find that 100 hours is sufficient to train a non-trivial, end-to-end speech system and perform meaningful experiments.

We apply the noise model trained on WSJ to the ground truth transcriptions in train-clean-100 in order to generate 18 noisy training datasets.
We will refer to each noisy dataset with the same name as the noise model used to generate it.
When Lead2Gold is used on a dataset, we use the corresponding noise model in equation \eqref{eq:l2g_numerator}.

\subsection{Evaluation}

To develop and test Lead2Gold we use the dev-clean and test-clean subsets of the LibriSpeech corpus.
We report the Word Error Rate (WER) between the decoded transcriptions produced by the ASR model and the ground truth transcriptions (without label noise) on these datasets. The WER on test-clean is reported both without and with language model (LM) rescoring. 

For the LM, we use a 4-gram trained on the LibriSpeech text training data with a 200~k word lexicon.
The hyperparameters of the decoder are tuned once on dev-clean with a model trained on the train-clean-100 subset with the ASG loss.
We use a beam size of 2,500, a beam threshold of 50, an LM weight of 2.66, a word insertion penalty of 1.33 and a space insertion penalty of -1.33.

The baseline is a model trained on the noisy dataset with the ASG loss.
We also train an oracle model on train-clean-100 without label noise using the ASG loss.

\subsection{Typical beam search results}

A typical beam search result is shown in Table~\ref{beam_res}. The \textit{Weight} column represents the contribution of a proposed transcription to the loss.
The weight for a given transcription is the logadd of the scores for all the alignments in the beam which lead to it.
We then normalize the weights with a softmax operation. 
The example shows that our method can recover good hypotheses and assign a reasonable weight to them. In this particular case, the best hypothesis is the ground truth.

\begin{table}
 
\resizebox{\columnwidth}{!}{
\noindent
\begin{tabular}{c|l|cc}
    \toprule
 & \multicolumn{1}{c|}{Transcription}  & Weight & LER\\ 
 \midrule
Ground truth  &   could not give his hand to the bride   & - & - \\
\midrule
$\tilde{Y}$ &  ool not ive his han to the rride   & - & 15.8 \\ 
\midrule
 & could not give his hand to the bride & 0.22 & 0 \\
 & could not ive his hand to the bride & 0.13 & 2.6 \\
 & could not give his hand to the brie & 0.05 & 2.6 \\
 & could not ive his hand to the brie & 0.03 & 5.3 \\
Transcription & coul not give his hand to the bride & 0.029 & 2.6 \\
hypotheses & ould not give his hand to the bride & 0.024 & 2.6 \\
 $Y^*$   & could not cive his hand to the bride & 0.023 & 2.6 \\            
 & could not give his and to the bride & 0.022 & 2.6 \\
 & coud not give his hand to the bride & 0.022 & 2.6 \\
 & could not give is hand to the bride & 0.018 & 2.6 \\
 & could not giv his hand to the bride & 0.018 & 2.6 \\
\bottomrule
\end{tabular}
} 
\caption{
    Example of the 10 best transcriptions $Y^*$ obtained by our beam search procedure. The Lead2Gold algorithm is applied on the ``M20 f1" noisy dataset with the corresponding noise model. We report the \textit{Weight} and the \textit{LER} between $Y^*$ and the ground truth. The noisy transcription is given by $\tilde{Y}$.
}
\label{beam_res}
\end{table}
\section{Experiments}
\label{sec:expes}

\subsection{Experimental protocol}
\label{ssec:protocol}

We implement the Lead2Gold (L2G) loss in the wav2letter++ framework \cite{pratap2018wav2letter++}, a fast speech recognition toolkit written in C++, and train a model on each of the 18 noisy datasets.
We first pre-train with the standard ASG loss without a noise model for 1,000 epochs.
This is done since the L2G loss is more than 15 times slower than ASG and can be up to 30 times slower when using a beam size of 300.
We then fine-tune with the L2G loss for 200 epochs in the substitution-only case and for 1 epoch in the general case (see explanation in Section \ref{sssec:cas2} below).
We train an oracle model on the clean data with the ASG loss for 1,200 epochs and during the last 200 epochs we anneal the learning rate by a factor of 2.

The network architecture is based on the off-the-shelf LibriSpeech ASG recipe provided with wav2letter++, namely, a 1D gated convolutional neural network (Gated ConvNet) \cite{liptchinsky2017based}.
The model contains 17 1D convolutional layers with a fixed kernel width of 13 followed by 2 linear layers.
For each layer, except for the last one, we apply weight normalization~\cite{salimans2016weight}, Gated Linear Units (GLUs)~\cite{dauphin2017language} as the activation function and a fixed dropout rate of 0.25.
Along the 17 convolutional layers we linearly increase the number of output hidden units from 100 to 200.
For the first layer we use a stride of 2 in order to reduce the number of frames used in computing L2G loss.
The first linear layer projects the number of hidden units to 400 and the last one to 29, producing one score for each output token.

The model contains only 10~M parameters, which is sufficient to accommodate the smaller training set.
The number of hidden units and hence parameters was tuned on train-clean-100 with the ASG loss using the dev-clean dataset for validation.
We do not perform further architecture search. Despite the fact that the overall architecture was designed for a larger dataset, 
we obtain a decent WER.
On dev-clean, the model achieves 16.9\% WER which is close to the 14.7\% WER reported for end-to-end models in \cite{luscher2019rwth}.
With an LM, the model reaches 9.5\% WER compared to 7.3\% WER from \cite{billa2017improving} which aims to find a good model in this setting.

The model takes 40 log-mel filterbank features as inputs.
We use the SGD optimizer and clip the norm of the gradients to 0.05. The learning rate is set to 8 when pre-training.
A separate learning rate of 0.004 is used to update the transition scores, $g(\pi_t|\pi_{t-1})$.

We train with a batch size of 320 and implement the L2G loss in parallel using one CPU per example in the batch.
We divide the loss by the square root of the length of the provided transcription.

During the pre-training phase we use 8 GPUs which gives an epoch time of approximately 30 seconds.
When training with the L2G loss, the number of GPUs used is not important since the computation of the loss, which occurs on the CPU, represents the bulk of the training time.
We set the beam size of L2G to 300 for all the experiments.
We find that increasing it further has little impact on the results.

The noise model weight $\alpha$ must be tuned.
Lead2Gold only converges with $\alpha$ below 0.7 in our experiments.
In general, as the amount of noise in the data the data grows, $\alpha$ must be shrinked in order to achieve convergence.
We reduce $\alpha$ until we obtain a WER convergence on dev-clean.

One epoch of L2G epoch lasts between 7 and 15 minutes.
In future work, we could reduce the complexity of Lead2Gold by using sub-word units or entire words as tokens.
This would allow us to use a smaller beam size and further reduce the frame rate of the encoded utterance.

\subsection{Results}
\label{sssec:cas2}

\begin{table}
\resizebox{\columnwidth}{!}{
\arrayrulecolor{black}
\begin{tabular}{lcccccc}
    \toprule
    ASR training & \multicolumn{2}{c}{dev-clean} & \multicolumn{2}{c}{test-clean} & \multicolumn{2}{c}{test-clean+LM}  \\
    dataset & ASG & L2G & ASG & L2G & ASG & L2G \\
   
   \midrule
   S M10 f0.5   & 17.5 & 17.1  & 17.7 & 17.3 & 12.1 & 10.2 \\
   S M10 f1     & 17.9 & 17.1 & 18.1 & 17.2 & 23.4 & 10.3 \\
   S M10 f2     & 18.8 & 17.4 & 18.8 & 18.6 & 21.4 & 10.4 \\
   \midrule
   S M20 f0.5   & 18.1 & 17.1 & 18.3 & 17.5 & 15.0 & 10.4 \\
   S M20 f1     & 19.0 & 17.5 & 19.2 & 18.1 & 22.3 & 10.7 \\
   S M20 f2     & 20.6 & 18.2 & 20.9 & 18.6 & 49.1 & 11.2 \\
   \midrule
   S M30 f0.5   & 18.3 & 17.3 & 18.8 & 17.6 & 19.9 & 10.6 \\
   S M30 f1     & 20.2 & 18.2 & 20.4 & 18.3 & 41.1 & 11.0 \\
   S M30 f2     & 24.8 & 19.6 & 25.1 & 20.0 & 80.1 & 12.7 \\
   \midrule
   clean-100h & 16.9 &  -  & 17.3 & - & 10.0 & - \\
   \bottomrule
\end{tabular}
}
\caption{
\label{tab:tab_subs}
WER (\%) achieved by ASR models trained on noisy data including substitutions only.}
\end{table}

Table \ref{tab:tab_subs} shows the results achieved by ASR models trained on noisy data including substitutions only.
We keep the learning rate to 8 and we set $\alpha$ to 0.5 except for the ``S M30 f2" case where it is set to 0.3.
Lead2Gold outperforms the ASG loss in every case.
We almost reach the oracle WER when the noise level is low, but the relative gain achieved by L2G is larger on noisier datasets.
For the model trained on the ``S M30 f2" dataset, we achieve a WER reduction of 5.1\% absolute when using the L2G loss.
We note that in some cases adding an external LM makes the WER worse when models are trained with the ASG loss on noisy data.
One possible explanation is that we did not re-tune the decoding parameters for each setting.

\begin{table}
\resizebox{\columnwidth}{!}{
\arrayrulecolor{black}
\begin{tabular}{lcccccc}
   \toprule
    ASR training & \multicolumn{2}{c}{dev-clean} & \multicolumn{2}{c}{test-clean} & \multicolumn{2}{c}{test-clean+LM}  \\
   dataset & ASG & L2G & ASG & L2G & ASG & L2G \\
   \midrule
   M10 f0.5   & 18.2 & 18.5 & 18.8 & 19 & 13.7 & 12.8 \\
   M10 f1     & 19.5 & 20.1 & 19.5 & 20.3 & 23.4 & 19.4 \\
   M10 f2     & 21.7 & 22.6 & 22.2 & 22.8 & 59.9 & 41.7 \\
   \midrule
   M20 f0.5   & 20.2 & 21.1 & 20.7 & 21.8 & 28.6 & 22.6 \\
   M20 f1     & 23.9 & 22.8 & 24.3 & 23.2 & 71.2 & 58.3 \\
   M20 f2     & 44.0 & 31.2 & 44.6 & 31.7 & 96.5 & 68.9 \\
   \midrule
   M30 f0.5   & 23.3 & 22.0 & 23.4 & 22.0 & 60.8 & 39.8 \\
   M30 f1     & 39.1 & 28.6 & 39.6 & 28.7 & 95.3 & 76.1 \\
   M30 f2     & 87.7 & 46.4 & 87.9 & 46.9 & 99.0 & 84.8 \\
   \midrule
   clean-100h & 16.9 &  - & 17.3 & - & 10.0 & - \\
   \bottomrule
\end{tabular}
}
\caption{
\label{tab:tab_general_case}
WER (\%) achieved by ASR models trained on noisy data including substitutions, insertions, and deletions.}
\end{table}

Table~\ref{tab:tab_general_case} shows the results achieved by ASR models trained on noisy data including substitutions, insertions, and deletions.
The learning rate is set to 1 and $\alpha$ to 0.1 for all of these experiments.
While L2G does not improve over the baseline ASG trained model  in every case, we still see a gain in performance in some cases, particularly at higher noise levels.
In the general case, we see more frequent convergence issues.
In some cases, the WER on the training data initially improves but then starts to deteriorate over time. We report the results when the best WER is reached on dev-clean during the single epoch we perform.
We found that if we continue to train, the predicted transcriptions tend to grow shorter and shorter.
One possible solution could be to use a different scale for each of the insertions, deletion and substitution probabilities.

\section{Conclusion}
\label{sec:conclu}

We propose Lead2Gold, a novel sequence-level loss function which incorporates a noise model and is able to better learn from noisy transcriptions. While the L2G objective is intractable and cannot be computed directly, we use a differentiable beam search to approximate it. We show that when contrived yet non-trivial noise is introduced into the labels used to train the acoustic model, L2G can dramatically outperform a noise-unaware criterion such as ASG.

The main limitation of Lead2Gold is the high computational cost.
In future work we plan to mitigate this problem in part by using sub-word units as tokens.
This would reduce the transcription length and thus make it possible to reduce the frame-rate of the encoded input utterance.
We also intend to further investigate scaling to many more CPUs and alternative SIMD style parallel implementations of the L2G loss-function.
Another avenue for improvement with L2G is using a more complex noise model.
The noise model can condition on the utterance or on some meta-data about the transcriber.

The L2G loss function has the potential to better leverage noisy data. Learning more accurately from noisy data can dramatically decrease the cost of generating transcriptions for acoustic training sets -- currently one of the biggest costs in developing a new speech recognition system.

\bibliographystyle{IEEEbib}
\bibliography{refs}

\begin{thebibliography}{10}

\bibitem{panayotov2015librispeech}
Vassil Panayotov, Guoguo Chen, Daniel Povey, and Sanjeev Khudanpur,
\newblock ``Librispeech: an {ASR} corpus based on public domain audio books,''
\newblock in {\em 2015 IEEE International Conference on Acoustics, Speech and
  Signal Processing (ICASSP)}. IEEE, 2015, pp. 5206--5210.

\bibitem{paul1992design}
Douglas~B Paul and Janet~M Baker,
\newblock ``The design for the {W}all {S}treet {J}ournal-based {CSR} corpus,''
\newblock in {\em Proceedings of the workshop on Speech and Natural Language}.
  Association for Computational Linguistics, 1992, pp. 357--362.

\bibitem{cieri2004fisher}
Christopher Cieri, David Miller, and Kevin Walker,
\newblock ``The fisher corpus: a resource for the next generations of
  speech-to-text.,''
\newblock in {\em LREC}, 2004, vol.~4, pp. 69--71.

\bibitem{godfrey1992switchboard}
John~J Godfrey, Edward~C Holliman, and Jane McDaniel,
\newblock ``Switchboard: Telephone speech corpus for research and
  development,''
\newblock in {\em [Proceedings] ICASSP-92: 1992 IEEE International Conference
  on Acoustics, Speech, and Signal Processing}. IEEE, 1992, vol.~1, pp.
  517--520.

\bibitem{vesely2013semi}
Karel Vesel{\`y}, Mirko Hannemann, and Luk{\'a}{\v{s}} Burget,
\newblock ``Semi-supervised training of deep neural networks,''
\newblock in {\em 2013 IEEE Workshop on Automatic Speech Recognition and
  Understanding}. IEEE, 2013, pp. 267--272.

\bibitem{karita2018semi}
Shigeki Karita, Shinji Watanabe, Tomoharu Iwata, Atsunori Ogawa, and Marc
  Delcroix,
\newblock ``Semi-supervised end-to-end speech recognition.,''
\newblock in {\em Interspeech}, 2018, pp. 2--6.

\bibitem{huang2016active}
Jiaji Huang, Rewon Child, Vinay Rao, Hairong Liu, Sanjeev Satheesh, and Adam
  Coates,
\newblock ``Active learning for speech recognition: the power of gradients,''
\newblock {\em arXiv preprint arXiv:1612.03226}, 2016.

\bibitem{drugman2019active}
Thomas Drugman, Janne Pylkkonen, and Reinhard Kneser,
\newblock ``Active and semi-supervised learning in {ASR}: Benefits on the
  acoustic and language models,''
\newblock {\em arXiv preprint arXiv:1903.02852}, 2019.

\bibitem{Yu2010active}
Dong Yu, Balakrishnan Varadarajan, li~Deng, and Alex Acero,
\newblock ``Active learning and semi-supervised learning for speech
  recognition: A unified framework using the global entropy reduction
  maximization criterion,''
\newblock {\em Computer Speech \& Language}, vol. 24, pp. 433--444, 2010.

\bibitem{wang2015transfer}
Dong Wang and Thomas~Fang Zheng,
\newblock ``Transfer learning for speech and language processing,''
\newblock in {\em Proc.\ APSIPA Annual Summit and Conf.}, 2015, pp. 1225--1237.

\bibitem{collobert2016wav2letter}
Ronan Collobert, Christian Puhrsch, and Gabriel Synnaeve,
\newblock ``Wav2letter: an end-to-end convnet-based speech recognition
  system,''
\newblock {\em arXiv preprint arXiv:1609.03193}, 2016.

\bibitem{patrini2017making}
Giorgio Patrini, Alessandro Rozza, Aditya Krishna~Menon, Richard Nock, and
  Lizhen Qu,
\newblock ``Making deep neural networks robust to label noise: A loss
  correction approach,''
\newblock in {\em Proceedings of the IEEE Conference on Computer Vision and
  Pattern Recognition}, 2017, pp. 1944--1952.

\bibitem{Label-Noise}
Takuhiro Kaneko, Yoshitaka Ushiku, and Tatsuya Harada,
\newblock ``Label-noise robust generative adversarial networks,''
\newblock {\em CoRR}, vol. abs/1811.11165, 2018.

\bibitem{sukhbaatar2014training}
Sainbayar Sukhbaatar, Joan Bruna, Manohar Paluri, Lubomir Bourdev, and Rob
  Fergus,
\newblock ``Training convolutional networks with noisy labels,''
\newblock {\em arXiv preprint arXiv:1406.2080}, 2014.

\bibitem{jindal2016learning}
Ishan Jindal, Matthew Nokleby, and Xuewen Chen,
\newblock ``Learning deep networks from noisy labels with dropout
  regularization,''
\newblock in {\em 2016 IEEE 16th International Conference on Data Mining
  (ICDM)}. IEEE, 2016, pp. 967--972.

\bibitem{goldberger2017training}
Jacob Goldberger and Ehud Ben-Reuven,
\newblock ``Training deep neural-networks using a noise adaptation layer,''
\newblock in {\em ICLR}, 2017.

\bibitem{xiao2015learning}
Tong Xiao, Tian Xia, Yi~Yang, Chang Huang, and Xiaogang Wang,
\newblock ``Learning from massive noisy labeled data for image
  classification,''
\newblock in {\em Proceedings of the IEEE conference on computer vision and
  pattern recognition}, 2015, pp. 2691--2699.

\bibitem{menon2016learning}
Aditya~Krishna Menon, Brendan Van~Rooyen, and Nagarajan Natarajan,
\newblock ``Learning from binary labels with instance-dependent corruption,''
\newblock {\em arXiv preprint arXiv:1605.00751}, 2016.

\bibitem{spellingCorrection}
Jinxi Guo, Tara~N Sainath, and Ron~J Weiss,
\newblock ``A spelling correction model for end-to-end speech recognition,''
\newblock in {\em ICASSP 2019-2019 IEEE International Conference on Acoustics,
  Speech and Signal Processing (ICASSP)}. IEEE, 2019, pp. 5651--5655.

\bibitem{hasegawa2017asr}
Mark~A Hasegawa-Johnson, Preethi Jyothi, Daniel McCloy, Majid Mirbagheri,
  Giovanni M~di Liberto, Amit Das, Bradley Ekin, Chunxi Liu, Vimal Manohar, Hao
  Tang, et~al.,
\newblock ``{ASR} for under-resourced languages from probabilistic
  transcription,''
\newblock {\em IEEE/ACM Transactions on Audio, Speech and Language Processing
  (TASLP)}, vol. 25, no. 1, pp. 50--63, 2017.

\bibitem{graves2006connectionist}
Alex Graves, Santiago Fern{\'a}ndez, Faustino Gomez, and J{\"u}rgen
  Schmidhuber,
\newblock ``Connectionist temporal classification: labelling unsegmented
  sequence data with recurrent neural networks,''
\newblock in {\em Proceedings of the 23rd international conference on Machine
  learning}. ACM, 2006, pp. 369--376.

\bibitem{povey2016purely}
Daniel Povey, Vijayaditya Peddinti, Daniel Galvez, Pegah Ghahremani, Vimal
  Manohar, Xingyu Na, Yiming Wang, and Sanjeev Khudanpur,
\newblock ``Purely sequence-trained neural networks for asr based on
  lattice-free mmi,''
\newblock in {\em Interspeech}, 2016, pp. 2751--2755.

\bibitem{collobert2019fully}
Ronan Collobert, Awni Hannun, and Gabriel Synnaeve,
\newblock ``A fully differentiable beam search decoder,''
\newblock {\em arXiv preprint arXiv:1902.06022}, 2019.

\bibitem{pratap2018wav2letter++}
Vineel Pratap, Awni Hannun, Qiantong Xu, Jeff Cai, Jacob Kahn, Gabriel
  Synnaeve, Vitaliy Liptchinsky, and Ronan Collobert,
\newblock ``Wav2letter++: The fastest open-source speech recognition system,''
\newblock {\em arXiv preprint arXiv:1812.07625}, 2018.

\bibitem{liptchinsky2017based}
Vitaliy Liptchinsky, Gabriel Synnaeve, and Ronan Collobert,
\newblock ``Letter-based speech recognition with gated convnets,''
\newblock {\em arXiv preprint arXiv:1712.09444}, 2017.

\bibitem{salimans2016weight}
Tim Salimans and Durk~P Kingma,
\newblock ``Weight normalization: A simple reparameterization to accelerate
  training of deep neural networks,''
\newblock in {\em Advances in Neural Information Processing Systems}, 2016, pp.
  901--909.

\bibitem{dauphin2017language}
Yann~N Dauphin, Angela Fan, Michael Auli, and David Grangier,
\newblock ``Language modeling with gated convolutional networks,''
\newblock in {\em Proceedings of the 34th International Conference on Machine
  Learning-Volume 70}. JMLR. org, 2017, pp. 933--941.

\bibitem{luscher2019rwth}
Christoph L{\"u}scher, Eugen Beck, Kazuki Irie, Markus Kitza, Wilfried Michel,
  Albert Zeyer, Ralf Schl{\"u}ter, and Hermann Ney,
\newblock ``{RWTH} {ASR} systems for librispeech: Hybrid vs attention-w/o data
  augmentation,''
\newblock {\em arXiv preprint arXiv:1905.03072}, 2019.

\bibitem{billa2017improving}
Jayadev Billa,
\newblock ``Improving {LSTM}-{CTC} based {ASR} performance in domains with
  limited training data,''
\newblock {\em arXiv preprint arXiv:1707.00722}, 2017.

\end{thebibliography}

\end{document}